\documentclass[letterpaper, 10 pt, conference]{ieeeconf}  
\usepackage{amsmath}  
\usepackage{graphicx}
\usepackage[caption=false,font=footnotesize]{subfig}  
\usepackage{xcolor}  
\usepackage{booktabs} 
\usepackage{bm}

\IEEEoverridecommandlockouts                              
\title{\LARGE \bf
Learn to Swim: Data-Driven LSTM Hydrodynamic Model for Quadruped Robot Gait Optimization}

\author{
    Fei Han$^{1,2}$, Pengming Guo$^2$, Hao Chen$^2$, Weikun Li$^2$,\\
    Jingbo Ren$^3$, Naijun Liu$^4$, Ning Yang$^4$, Dixia Fan$^{2 *}$ \\
    \thanks{* Corresponding author: Dixia Fan}
    \thanks{$^{1}$ Zhejiang University, Hangzhou, 310027, China.}
    \thanks{$^{2}$ FSI lab, Westlake University, Hangzhou, 310030, China.}
    \thanks{$^{3}$ Xinyang Normal University, Xinyang, 464000, China}
    \thanks{$^{4}$ Institute of Automation, Chinese Academy of Sciences, Beijing, 100190, China.}
    \thanks{E-mails: fandixia@westlake.edu.cn, hanfei@westlake.edu.cn}
    \thanks{Notice: This work has been accepted for publication in the IEEE International Conference on Robotics and Automation (ICRA) 2025. The final version will be available in IEEE Xplore (DOI to be assigned upon publication).}   }
\begin{document}

\maketitle
\thispagestyle{empty}
\pagestyle{empty}

\begin{abstract}
This paper presents a Long Short-Term Memory network-based Fluid Experiment Data-Driven model (FED-LSTM) for predicting unsteady, nonlinear hydrodynamic forces on the underwater quadruped robot we constructed. Trained on experimental data from leg force and body drag tests conducted in both a recirculating water tank and a towing tank, FED-LSTM outperforms traditional Empirical Formulas (EF) commonly used for flow prediction over flat surfaces. The model demonstrates superior accuracy and adaptability in capturing complex fluid dynamics, particularly in straight-line and turning-gait optimizations via the NSGA-II algorithm. FED-LSTM reduces deflection errors during straight-line swimming and improves turn times without increasing the turning radius. Hardware experiments further validate the model's precision and stability over EF. This approach provides a robust framework for enhancing the swimming performance of legged robots, laying the groundwork for future advances in underwater robotic locomotion.

\end{abstract}

\section{INTRODUCTION}
Legged robots outperform wheeled robots in rough environments due to their flexibility and ability to cross obstacles, making them ideal for tasks like search and rescue \cite{biswal2021development}. Recent advancements in artificial intelligence, particularly in control algorithms, have significantly improved quadrupedal robots, enhancing their performance in dynamic, unstructured settings \cite{he2020mechanism}.

Various motion control strategies have been developed, including gait planning based on the Central Pattern Generators (CPG) algorithm \cite{wei2022cpg, zhong2018cpg}, Adaptive Variable Impedance Control (AVIC) for unknown terrains \cite{xu2020adaptive}, Model Predictive Control (MPC) leveraging body dynamic models \cite{kwon2020fast, hwangbo2019learning}, and Reinforcement learning-based algorithms \cite{lee2021reinforcement, aractingi2023controlling}, all aiming to optimize performance across diverse environments.

\begin{figure}[htbp]
    \centering
    {\includegraphics[width=0.9\linewidth]{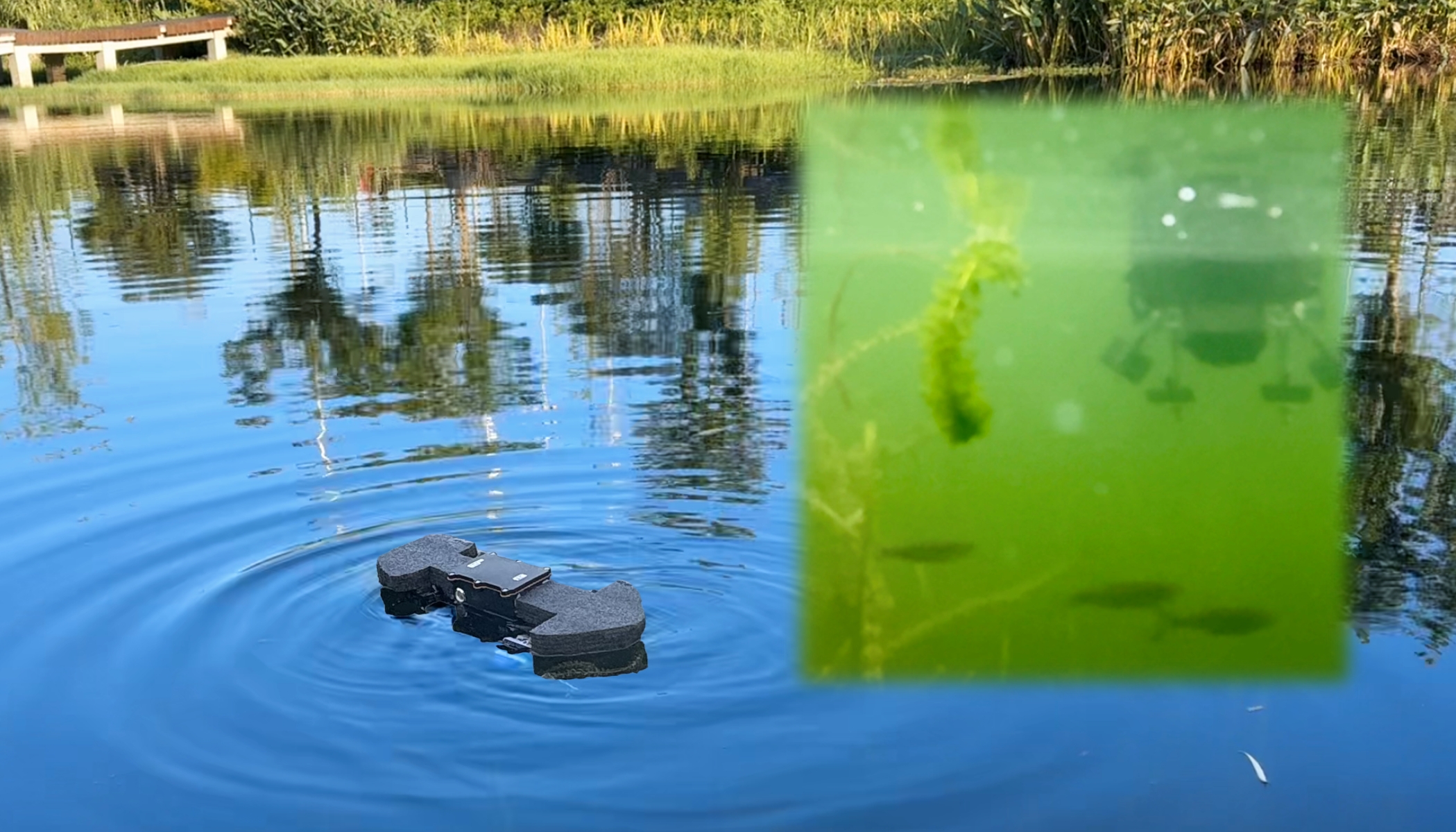}}
    \caption{
        The demonstration of the swimming quadruped robot naturally integrating into the aquatic environment without disturbing the surrounding fish and plant.
    }
    \label{fig:example}
\end{figure}

However, while legged robots excel on land, they face significant challenges in aquatic environments where control strategies differ drastically \cite{chignoli2021online, chen2018controllable}. In nature, amphibians transition seamlessly between land and water, adapting to static terrestrial and dynamic aquatic environments. Aquatic settings involve continuous perturbations from water viscosity \cite{park2008dynamic}, making land-based control strategies ineffective \cite{zufferey2022between}. Hydrodynamic challenges, such as boundary layers and vortex separation, introduce complex fluid resistance that traditional control models fail to address \cite{wang2020development}.

Accurate hydrodynamic modeling is essential for controlling legged robots as they move in water. Without accurate force measurements and detailed fluid dynamic analysis, control strategies remain limited. Previous methods, relying on mechanical adjustments \cite{vogel2014design, baines2019toward, ren2021research} and sample-averaged force models \cite{yao2023learning}, lack the precision needed to optimize robot performance. Addressing this gap requires real-world fluid force data, making the optimization of aquatic robots a complex and nontrivial challenge.

\begin{figure*}[t]
    \centering
    \includegraphics[width=\linewidth]{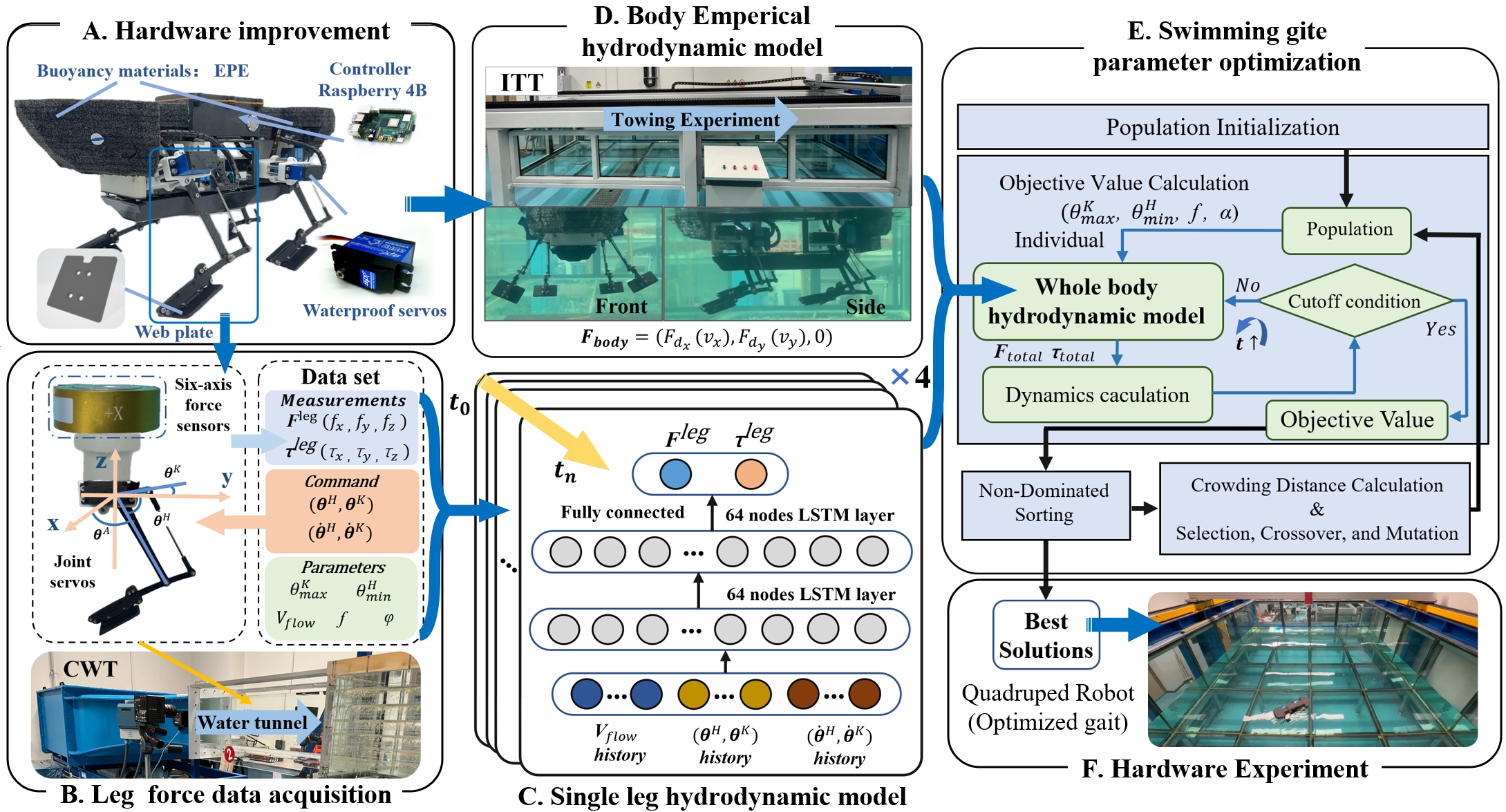} 
    \caption{The Scheme of the Robot Hydrodynamic Testing and Optimization. Panel A shows the overview of the swimming quadruped robot. Panel B shows the hydrodynamic experiment of the leg peddling in the water tunnel. Panel C depicts the EFD-LSTM model's structure, inputs, and outputs. Panel D reports the towing test for determining body hydrodynamics. Panel E highlights the key process of the robot swimming gait optimization, which is deployed in the open water test, shown in Panel F.}
    \label{workflow}
\end{figure*}

In this research, we propose an advanced approach to improve the water movement ability of quadruped robots. Our contributions include:
\begin{enumerate}
    \item \textbf{FED-LSTM Model for Hydrodynamics:} A fluid experimental data-driven Long Short-Term Memory (FED-LSTM) model that captures complex hydrodynamics on each leg, combined with an experimentally derived body fluid model.
    \item \textbf{Gait Optimization with NSGA-II:} Optimization of swimming gait parameters using the NSGA-II multi-objective algorithm, improving underwater efficiency and maneuverability.
    \item \textbf{Hardware Enhancements:} Waterproofing and the addition of a rigid web to the 12-degree-of-freedom Pupper locomotion system, mimicking aquatic animal movements for more effective swimming.
    \item \textbf{Experimental Validation:} Real-world experiments demonstrating significant improvements in swimming performance and control in complex water environments.
\end{enumerate}

These contributions provide a robust framework for advancing hydrodynamic modeling and gait planning of legged aquatic robots.

\section{SYSTEM OVERVIEW}

Our system stems on the Pupper quadruped robot \cite{pupper}, which has a symmetric design with three joints per leg, shown in Fig.~\ref{workflow}B. The joint angle of Hip Abduction / Adduction (HAA) $\theta^A$, responsible for lateral movement, is driven by a single servo motor. The joint angle of Hip Flexion / Extension (HFE) $\theta^H$ and the joint angle of Knee Flexion / Extension (KFE) $\theta^K$ control the swing of the thigh and calf, a quadrilateral linkage mechanism couples these two joints, allowing coordinated leg adjustments using two servo motors.

We modify the robot by adding a rigid square web ($36\, cm^2$, $3\, mm$ thick) to the midsection of each leg (Fig.~\ref{workflow}A). This modification allows significant production of hydrodynamic forces without interfering with terrestrial movement. 

For our initial underwater experiments, we add Expanded Polyethylene (EPE) buoyant materials to give the robot positive buoyancy, ensuring stability while swimming. The robot is leveled to prevent trimming (longitudinal tilt) and heel (lateral tilt). To minimize drag, the buoyant materials at the head and tail are shaped to resemble streamlined forms, inspired by boats and sharks.

The Pupper robot is equipped with a Raspberry Pi 4B, and we upgrad the 12 waterproof servo motors with a peak torque of $3.5\, N\cdot m$, providing sufficient torque to overcome the robot's large drag underwater. A custom-built waterproof compartment protects the controller and battery, ensuring reliable operation in water.

\section{Robot Hydrodynamic Model construction}
To address the complex fluid dynamics during the robot paddling \cite{xue2023exploring}, we develop a hydrodynamic model by decomposing it into two parts: the leg and the body.
\begin{figure}[ht]
    \centering
    \includegraphics[width=0.8\linewidth]{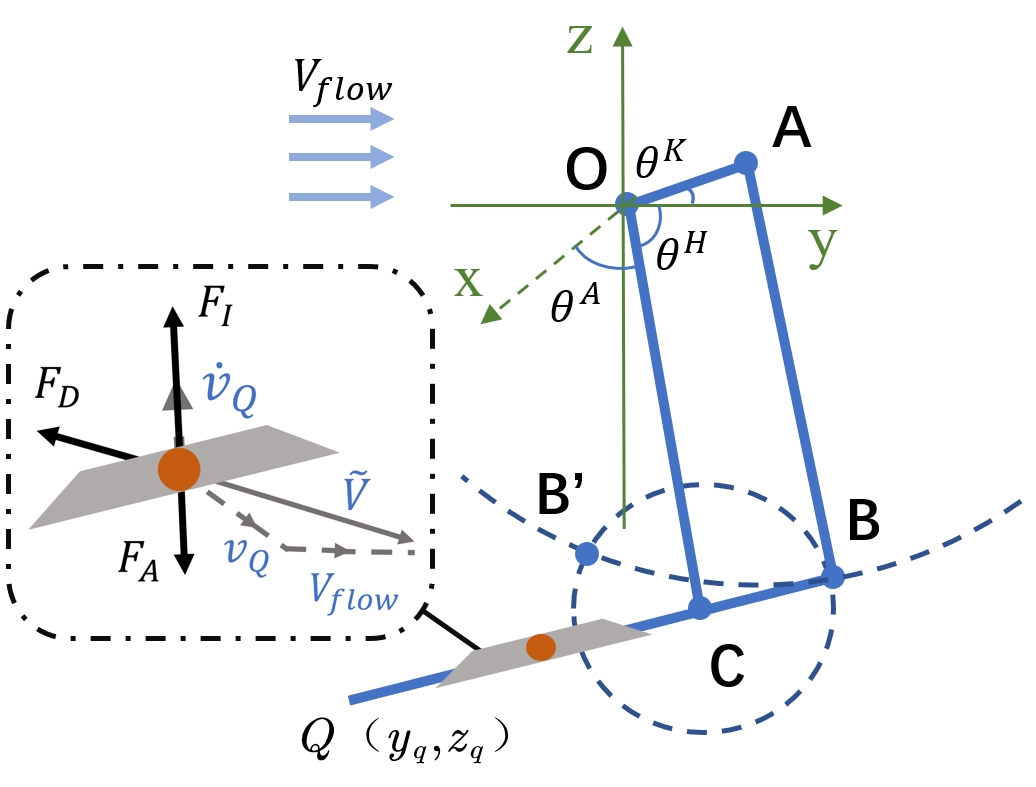} 
    \caption{Definition of leg kinematics and fluid force model: The trajectory of the web midpoint Q is calculated from $\theta_H$ and $\theta_K$, and is used to compute the empirical force components. The force direction corresponding to the motion and fluid direction is highlighted in the dashed box. }
    \label{fig:exampleq}
\end{figure}
\subsection{Empirical Hydrodynamic Modeling of Leg Paddling}

\subsubsection{Leg Kinematics}

The planar motion of the rigid web can be modeled as a 2D slender body in uniform flow, shown in Fig.~\ref{fig:exampleq}. The orientation and trajectory of the web midpoint Q is calculated based on the joint angles $\theta^H$ and $\theta^K$ using the geometry of points A, B, and C, given $OA = BC = 0.035\,m$, $OC = AB = 0.125\,m$, and BQ is $2.5$ times the length of $BC$.

Given a set of angles $\theta^H$ and $\theta^K$, the coordinates of points $A$ and $C$ can be expressed. Point $B$, with the larger $Y$-coordinate, is determined using points $A$ and $C$ as circle centers and $AO$ and $CB$ as radii. The coordinates of point $Q$ are then derived from points $B$ and $C$ based on the linkage geometry.
\subsubsection{Fluid Force}
Based on the quasi-steady assumption, the fluid force $F_{\text{T}}$ acting on the center of the web Q can be calculated as follows:
\begin{equation}
    F_T = F_A + F_D + F_I,
\end{equation}
where the added mass force $F_A$, the drag force $F_D$ and the inertial force of the web $F_I$ can be expressed as follows:

\begin{equation}
    \left\{
    \begin{split}
        F_A &= 2\pi \rho_{\text{water}} a^3 \cdot \dot{v}_{Q},\\
        F_D &= 0.5 \rho_{\text{water}} S C_R |\tilde{V}| \tilde{V},\\
        F_I &= m_{\text{web}} \cdot \dot{v}_{Q}, \\
    \end{split}
    \right.
\end{equation}
where $a$ is the characteristic length, $S = (2a)^2$ is the reference area, and $v_{Q}$ is the normal component velocity at point Q. Based on the shape of one leg and its motion, the drag coefficient $C_R$ is chosen as $C_R = 0.7$\cite{hoerner1965fluid}, the mass $m_{\text{web}} = 10$g.

\subsection{FED-LSTM Model of Leg}

\subsubsection{Fluid Force Acquisition}
As shown in Fig. \ref{workflow}B, experiments of leg paddling are conducted in a water tunnel, where a six-component DMI D6095XA transducer is used to measure the instantaneous forces and torques acting on the robotic leg. The prescribed motion of the leg is governed by the following equations:

\footnotesize 
\begin{equation}
\left\{
\begin{split}
    \theta^H(t) &= \frac{\theta^H_{max} - \theta^H_{min}}{2} \sin(2\pi f t+\alpha) + \frac{\theta^H_{max} + \theta^H_{min}}{2},\\
    \theta^K(t) &= \frac{\theta^K_{max} - \theta^K_{min}}{2} \sin(2\pi f t+\phi+\alpha) + \frac{\theta^K_{max} + \theta^K_{min}}{2},
\end{split}
\right.
\label{brute_force}
\end{equation}
\normalsize
where, \(\theta^H_{\text{max}}\) and \(\theta^H_{\text{min}}\) represent the maximum and minimum angles of the HFE joint, while \(\theta^K_{\text{max}}\) and \(\theta^K_{\text{min}}\) represent those of the KFE joint. The motion frequency is denoted by \(f\), and \(\phi\) is the phase difference between the HFE and KFE joints. For this analysis, \(\theta^H_{\text{max}} = -100^\circ\) and \(\theta^K_{\text{min}} = 80^\circ\) are fixed, and by varying \(\theta^H_{\text{min}}\) and \(\theta^K_{\text{max}}\), the joint motion range is adjusted. In this part, \(\alpha\) is set to zero.

Table \ref{tab:parameter_ranges1} lists the test ranges of these parameters. Data are collected by matching the parameters listed in the table, capturing 10 cycles of motion and force data for each parameter set at a given motion frequency $f$. A total of 480 sets of parameters are tested for each flow speed, with a sampling frequency of $65\,Hz$, resulting in approximately 3 million data points covering the relationship between control angles, control angular velocities, and the corresponding dynamic data.

\begin{table}[htbp]
\caption{Experimental Parameter Ranges}
\begin{center}
\renewcommand{\arraystretch}{1.4} 
\resizebox{\linewidth}{!}{
\begin{tabular}{|c|c|c|c|c|}
\hline
\textbf{Parameter}  & $\theta^H_{\text{min}}(^\circ)$ & $\theta^K_{\text{max}}(^\circ)$ & $f(Hz)$ & $\phi$ \\
\hline
\textbf{Range} & $[10, -50]$ & $[-20, -80]$ & $[0.3, 0.6]$ & $[{\pi}/3, {5\pi}/3]$ \\
\hline
\textbf{Interval} & $20$ & $20$ & 0.1 & $\pi/3$ \\
\hline
\end{tabular}
}
\label{tab:parameter_ranges1}
\end{center}
\end{table}

\vspace{-2mm}  
\subsubsection{Leg Hydrodynamic Model Training}
To model the nonlinear and unsteady hydrodynamics of the paddling leg, we deploy a Long Short-Term Memory (LSTM) network, trained using fluid experiment data. The FED-LSTM model, depicted in Fig.~\ref{workflow}C, comprises two LSTM layers, each with 64 hidden units, and a dropout rate of 0.21. The hidden states from the final LSTM time step are passed through a fully connected layer to predict hydrodynamic forces and torques. The learning rate is adaptively adjusted, ranging between 0.001 and 0.1.

To improve sensitivity to flow velocity, quadratic interpolation is applied to the velocity data at 0.05, 0.15 and $0.25\,m/s$, based on the proportional relationship between velocity and force. After interpolation, low-pass filtering is performed to maintain consistency. The dataset is divided into training, validation, and test sets in a 7:1:2 ratio. Validation loss is monitored after each training epoch to adjust the learning rate dynamically, with dropout applied to prevent overfitting.

Low-pass filtering (cutoff frequency of $6\, Hz$) is used in preprocessing to remove high-frequency noise while preserving higher-order terms in the fluid force data. The input sequence to the model consists of 16-time steps, each containing the tuple $(V_{\text{flow}}, \theta^H, \theta^K, \dot{\theta}^H, \dot{\theta}^K)$, representing flow velocity, joint angles, and angular velocities. The model outputs three torques and three forces $(\tau^{\text{leg}} (\tau_x, \tau_y, \tau_z), F^{\text{leg}} (f_x, f_y, f_z))$ acting on the leg.

\subsection{Comparison Between FED-LSTM And EF Model}

\begin{figure}[ht]
    \centering
    \includegraphics[width=\linewidth]{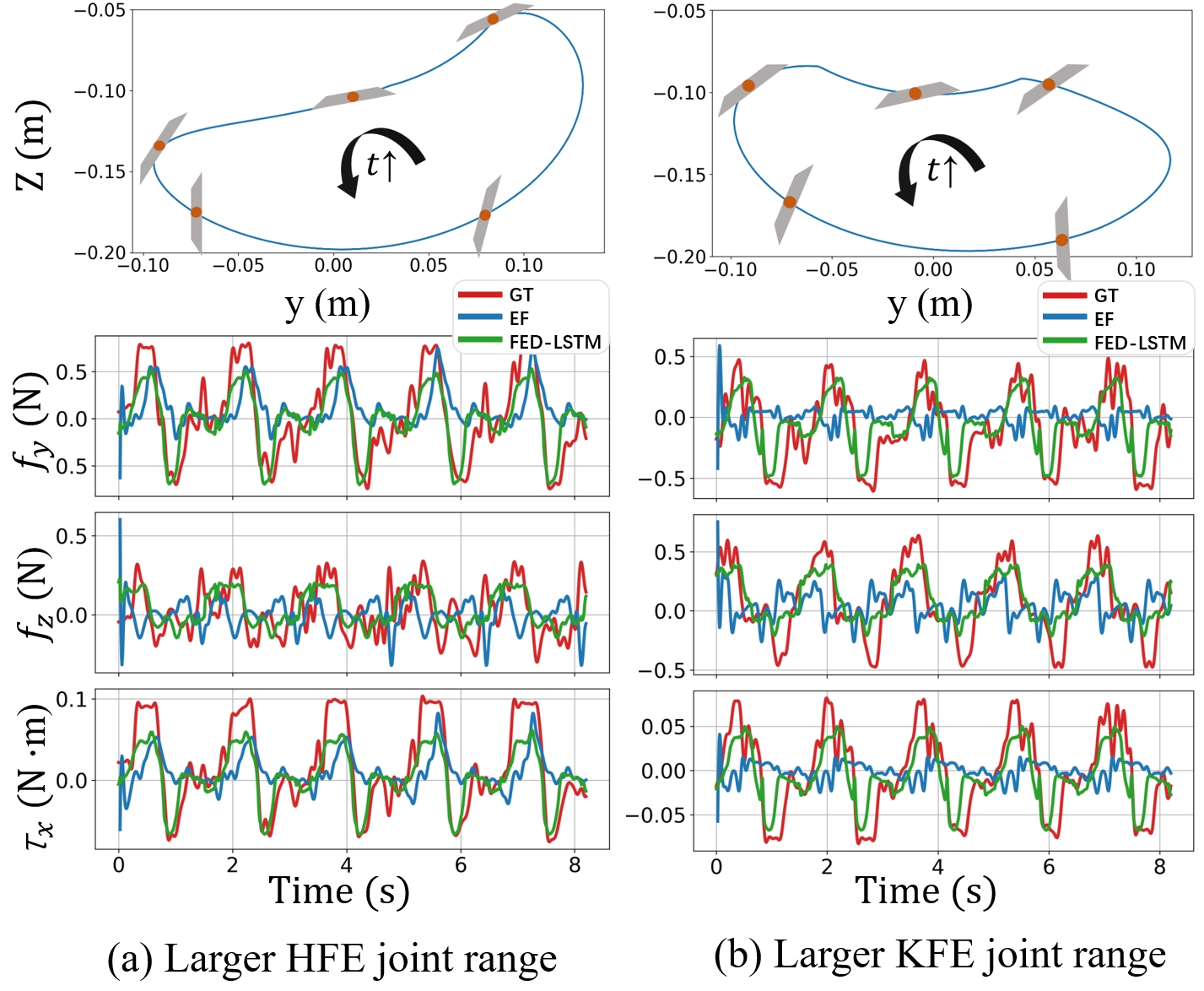} 
    \caption{Comparison of FED-LSTM and EF Models for Force and Torque Prediction: The first row illustrates the trajectories and the orientation of the web. The comparison among the measured force and torque as the ground truth (GT, red) and two models' predictions highlights that FED-LSTM (green) performs significantly better than the EF (blue) model in capturing the complex hydrodynamics of the paddling leg.}
    \label{prediction curve}
\end{figure}

\begin{figure}[ht]
    \centering
    \includegraphics[width=0.9\linewidth]{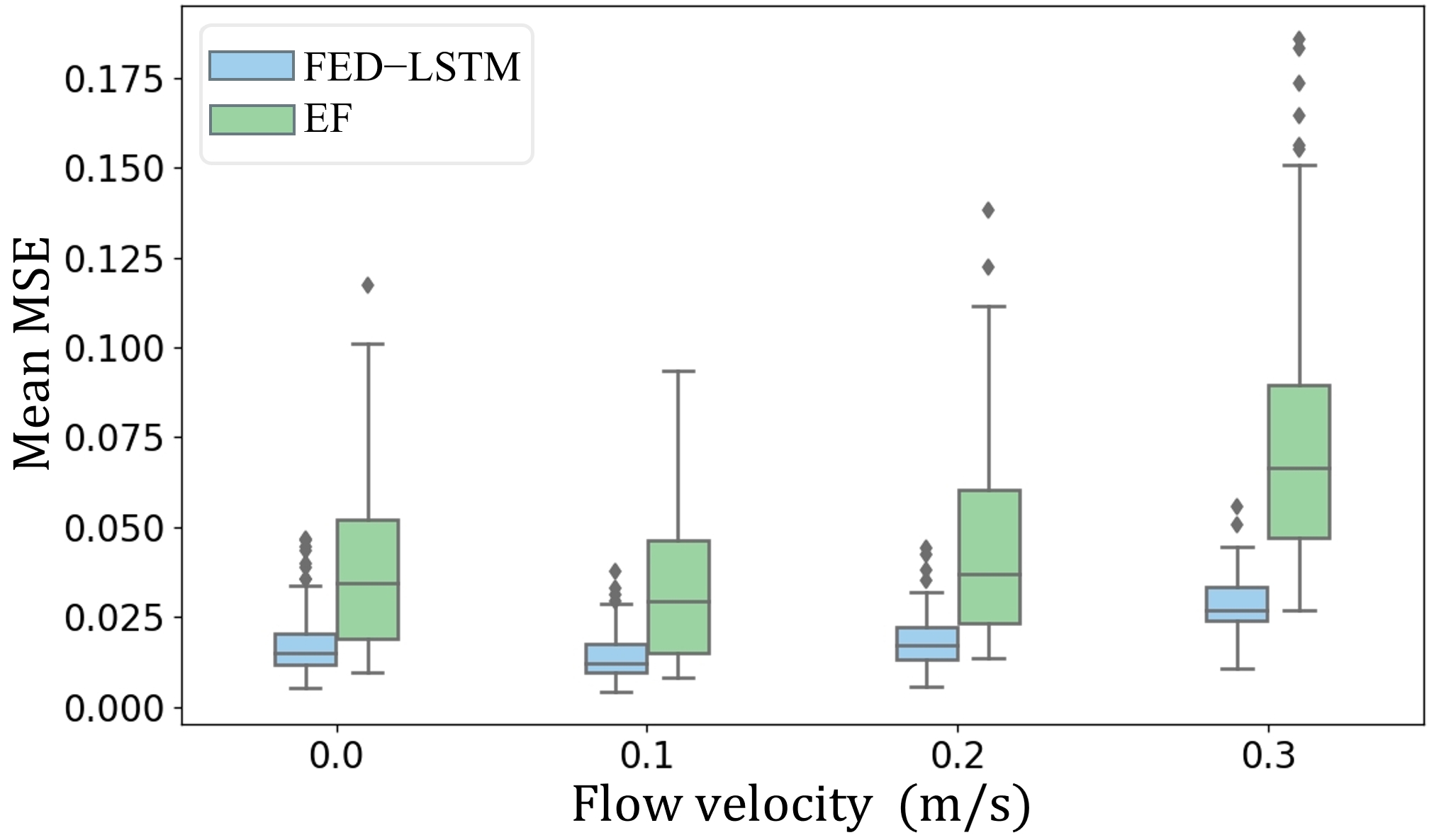} 
    \caption{Comparison of Prediction MSE between FED-LSTM and EF Models: Each individual data point represents the mean of MSE of $f_y, f_z$ and $\tau_x$  for a specific motion parameter at a given flow velocity. Box represents the range of MSE of model predictions at a given flow speed. The central line within the box denotes the median, while the box's edges indicate the 25$^{th}$ and 75$^{th}$ percentiles of the error distribution. The whiskers extend to 1.5 times the interquartile range, capturing the minimum and maximum values of non-outlier data. Outliers, which indicate abnormal errors for specific motion parameters, are represented by individual points outside the whiskers.}
    \label{fig_boxplot}
\end{figure}

As shown in Fig.~\ref{prediction curve}(a) for $\theta^H_{\text{min}} = 10^\circ$ and $\theta^K_{\text{max}} = -20^\circ$, and Fig.~\ref{prediction curve}(b) for $\theta^H_{\text{min}} = -50^\circ$ and $\theta^K_{\text{max}} = -60^\circ$, while the EF model captures qualitatively the trend of the measured force profile, the FED-LSTM model significantly outperforms it with lower Mean Squared Errors (MSE). Furthermore, the EF model performs poorly in predicting the torque of the paddling leg. This indicates that the EF model works reasonably under simpler conditions but struggles with more complex motion patterns, whereas the FED-LSTM model remains robust.
To further validate the model's performance, we plot in Fig.~\ref{fig_boxplot} the MSE comparison between two models for different flow speeds from 0 to $0.3\,m/s$. As the flow velocity increases, the mean square error (MSE) of both models also rises; however, the errors of the EF model increase more sharply. At $0.2\,m/s$ and $0.3\,m/s$, the EF model's MSE expands considerably, indicating higher uncertainty and deviation from the ground truth. In contrast, the FED-LSTM model maintains consistently low and stable MSE across all flow velocities. In other words, the FED-LSTM model proves to be more robust in challenging scenarios, capturing the nonlinear and unsteady nature of the peddling leg.

\subsection{Body Hydrodynamic Modeling}

The forward drag (in the direction of motion) and the lateral drag (perpendicular to the direction of motion) of the robot are measured through towing experiments in a 4-meter-long towing tank. The robot is towed from 0.1 to $0.35\,m/s$, and the result shows that the drag can be modeled in a quadratic form as follows:

\begin{equation}
    \left\{
    \begin{split}
        F_{d_y} &= 9.997 v_y^2 - 0.132 v_y + 0.334,\\
        F_{d_x} &= 15.571 v_x^2 + 0.937 v_x + 0.055, \\
    \end{split}
    \right.
\end{equation}
where \( F_{d_y} \) and \( F_{d_x} \) are the forward and lateral drag, where \( v_y \) and \( v_x \) are the forward and lateral speed. 

\subsection{Full-Body Dynamics}

The full-body dynamics of the robot incorporates both the forces and torques acting on the HAA joints' angle \( \bm{\theta}^{\text{A}} \) of each leg and the drag force on the body in the fluid as follows: 
\begin{equation}
    (\bm{F/\tau})_{\text{total}} = \sum_{i=1}^{4} (\bm{F/\tau})^{\text{Mc}}_i + (\bm{F/\tau})_{\text{body}},
\end{equation}
where the total force or torque \( (\bm{F/\tau})_{\text{total}} \) on the system includes the contributions from the drag force acting on the robot's body \( (\bm{F/\tau})_{\text{body}} \), each leg paddling force and torque to the robot's metacenter $(\bm{F/\tau})^{\text{Mc}}_i$, and \( i = \{\text{LF}, \text{RF}, \text{LH}, \text{RH}\} \). 

In particular, each leg rotates around the axis \( \bm{R} = [r_x, 0, r_z] \) by an angle \( \theta^{\text{A}} \) at point \( \bm{p}^{\text{leg}} = [x^{\text{leg}}, y^{\text{leg}}, z^{\text{leg}}] \) from the metacenter as the origin. Therefore, the force of each leg $(\bm{F})^{\text{Mc}}_i$ on the metacenter can be calculated as follows: 
\begin{equation}
    \bm{F}^{\text{Mc}} = R(\bm{R}, \theta^{\text{A}}) \bm{F}^{\text{leg}},
\end{equation}
where $\bm{F}^{\text{leg}}$ is the force of each leg acting on the joint, namely measured by the force sensor, the rotation matrix is $R(\bm{R}, \theta^{\text{A}}) = I + \sin(\theta^{\text{A}}) \bm{K}_{\bm{R}} + (1 - \cos(\theta^{\text{A}})) \bm{K}_{\bm{R}}^2$,  \( I \) is the identity matrix, and \( \bm{K}_{\bm{R}} \) is the skew-symmetric matrix derived from the rotation axis \( \bm{R} \). Then the torque at the metacenter can be computed as follows: 
\begin{equation}
    \bm{\tau}^{\text{Mc}} = \bm{p}^{\text{leg}} \times \bm{F}^{\text{Mc}} + R(\bm{R}, \theta^{\text{A}}) \bm{\tau}^{\text{leg}}.
\end{equation}

\section{Motion Parameter Optimization - Gait Optimization}

In this section, we describe the initial conditions of the quadruped robot and introduce the optimization parameters. The angle of the front leg \(\theta^A\) is set to 30° (upward from the horizontal), while the HFE and KFE servos of the hind legs remain horizontal to avoid collisions. Then, the swimming performance of the quadruped robot is influenced by the motor angles of the HFE and KFE joints, which are controlled by sine functions. The key optimization parameters include the maximum amplitude of HFE ($\theta^H_{\text{max}}$), the minimum amplitude of KFE ($\theta^K_{\text{min}}$), the movement frequency $f$, and the phase difference $\alpha$ among the four legs.

The joint angles for each leg are same as Equation~\ref{brute_force}, where \(\theta^H_{\text{max}} = -100^\circ\) and \(\theta^K_{\text{min}} = 80^\circ\) are fixed, and the phase $\phi$ between $\theta^H$ and $\theta^K$ is set to be $\pi/3$. In addition, by adding a phase difference $\alpha$ to each leg, we create phase differences among the four legs, thereby optimizing their coordination. Therefore, the ranges for the parameters are set as follows:

\begin{table}[htbp]
\renewcommand{\arraystretch}{1.5}  

\setlength{\tabcolsep}{8pt}  
\caption{Parameter Ranges for Gait Optimization}
\begin{center}
\resizebox{\linewidth}{!}{
\begin{tabular}{|c|c|c|c|c|}
\hline
\textbf{Parameter} & $\theta^H_{\text{min}}(^\circ)$ & $\theta^K_{\text{max}}(^\circ)$ & $f(Hz)$ & $\alpha_i$ \\
\hline
\textbf{Range} & $[10, -50]$ & $[-20, -80]$ & $[0.2, 0.65]$& $[0, 2\pi]$ \\
\hline
\end{tabular}
}
\label{tab:parameter_ranges}
\end{center}
\end{table}

\subsection{Dynamic Parameter Calculation}

The swimming motion dynamics are computed on the basis of the robot's total forces and torques. The yaw acceleration, velocity, and displacement are calculated using:
\begin{equation}
    \ddot{\theta}_{\text{yaw}} = \frac{\tau_z}{I_{\text{yaw}}}, \quad
    \dot{\theta}_{\text{yaw}}(t + \Delta t) = \dot{\theta}_{\text{yaw}}(t) + \ddot{\theta}_{\text{yaw}} \Delta t,
\end{equation}
\begin{equation}
    \theta_{\text{yaw}}(t + \Delta t) = \theta_{\text{yaw}}(t) + \dot{\theta}_{\text{yaw}}(t) \Delta t,
\end{equation}

For translational motion in the $x$ and $y$ directions:
\begin{equation}
    (\ddot{x}, \ddot{y}, 0) = \frac{F_{\text{total}}}{m},
\end{equation}
\begin{equation}
    (\dot{x}/\dot{y})(t + \Delta t) = (\dot{x}/\dot{y})(t) + (\ddot{x}/\ddot{y}) \Delta t,
\end{equation}
\begin{equation}
\left\{
    \begin{aligned}
    x_{\text{global}}(t + \Delta t) &= x_{\text{global}}(t) + \dot{x}(t) \Delta t, \\
    y_{\text{global}}(t + \Delta t) &= y_{\text{global}}(t) + \dot{y}(t) \Delta t,
    \end{aligned}
\right.
\end{equation}
where $y_{\text{global}}$ and $x_{\text{global}}$ are the coordinates in the world coordinate system with the starting point of the movement as the origin.

\subsection{Objective Functions for Motion Optimization}

\begin{table}[htbp]
\centering
\renewcommand{\arraystretch}{1.6}  
\setlength{\tabcolsep}{8pt}  
\caption{Objective Functions for Two Motions}
\label{tab:motion_objectives}
\begin{tabular}{|c|c|}
\hline
\textbf{Physical Objective}                & \textbf{Objective Function} \\ \hline

\multicolumn{2}{|c|}{\textbf{Straight-Line Motion}} \\ \hline

Y-axis impulse                 & $f_1 = -\sum_{i=0}^{N-1} F_{\text{total}_y}(t_i) \Delta t$ \\ \hline
Yaw angle error                & $f_2 = |\theta_{\text{yaw}}(t_{\text{final}}) - \theta_{\text{target}}|$ \\ \hline
Runtime                        & $f_3 = t_{\text{final}}$ \\ \hline

\multicolumn{2}{|c|}{\textbf{Turn Motion}} \\ \hline

Path length                    & $f_1 = \sum_{i=0}^{N-1} \sqrt{\Delta x_{\text{global}}(t_i)^2 + \Delta y_{\text{global}}(t_i)^2}$ \\ \hline
Yaw angle error                & $f_2 = |\theta_{\text{yaw}}(t_{\text{final}}) - \theta_{\text{target}}|$ \\ \hline
Runtime                        & $f_3 = t_{\text{final}}$ \\ \hline
\end{tabular}
\end{table}

In the NSGA-II algorithm, the objective is to minimize the optimization targets. An evaluation metric $S$ is introduced to assess the quality of the solutions, with the main motion objectives (straight-line and turning) weighted accordingly. The evaluation score for each solution is given as follows:
\begin{equation}
    S = w_1 \cdot f_1 + w_2 \cdot f_2 + w_3 \cdot f_3,
\end{equation}
where the weights are $w_1 = 1$, $w_2 = 4$, and $w_3 = 2$, corresponding to the objective functions $f_1$, $f_2$, and $f_3$, respectively listed in Table \ref{tab:motion_objectives}. The solutions are ranked based on the score $S$, with lower scores indicating better solutions. The best 8 solutions are retained for hardware experiments.

\section{Open Water Test with Optimized Gaits}

\begin{figure}[ht]
    \centering
    \includegraphics[width=\linewidth]{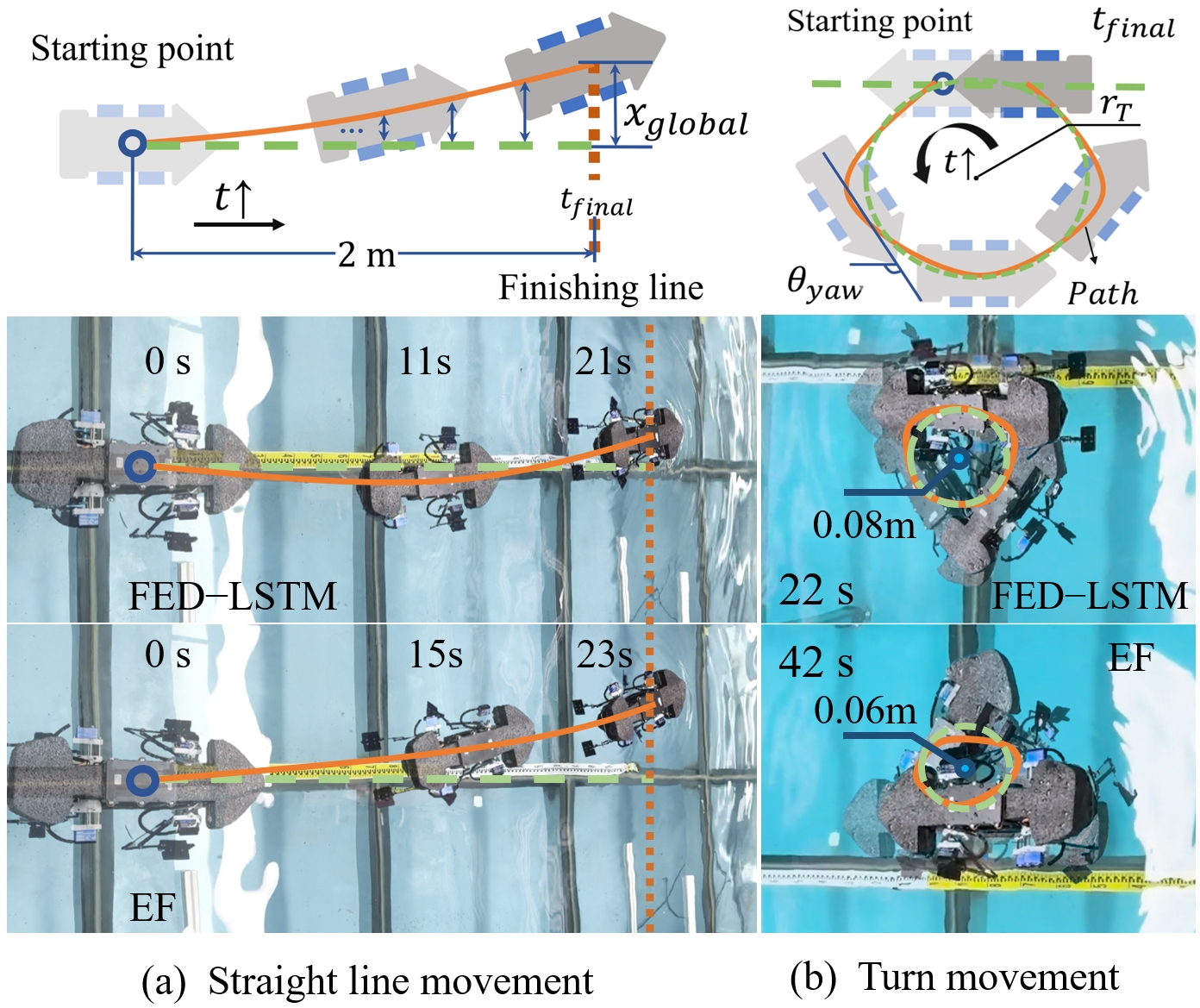} 
    \caption{The Robot Open Water Test: The comparison of optimized swimming gait using FED-LSTM ($2^{nd}$ row) and EF ($3^{rd}$ row) model for straight swimming (a) and turning (b). The result shows a significant improvement in speed, accuracy, and turning radius when using the FED-LSTM model.}
    \label{HardwarePerformance}
\end{figure}

We evaluate the robot’s hardware performance in the open water test using the best solutions from both the FED-LSTM and EF models. The robot’s movement is tested in a $4\,m \times 5\,m \times 1.5\,m$ water tank for both straight-line and turning movements. In Fig.~\ref{HardwarePerformance}, we compare the smallest cost $S$, representing the numerically best solution, for both models.

When moving straight, the robot is considered to finish the task when either reaching the finishing line or spending more than $60\,s$. We measured the global $X_{\text{global}}$ position between 0 and $2\,m$ at intervals of $0.25\,m$, calculating the Mean Absolute Error of $X_{\text{global}}$ (MAE($X_{\text{global}}$)) for each best solution. The FED-LSTM model completed the task in $21\,s$ with MAE($X_{\text{global}}$) of $0.09\,m$, while the EF model took $23\,s$ with MAE($X_{\text{global}}$) of $0.17\,m$. This represents a 47.1\% improvement in positional accuracy and an 8.7\% reduction in completion time for the FED-LSTM model.

In the turning movement task, starting from an initial deflection angle, the robot either completes a 360° turn or stops after $60\,s$. The trajectory is fitted to a circle, and the turning radius and completion time are recorded. The FED-LSTM model completes the turn in $22\,s$ with a fitted radius of $0.08\,m$, while the EF model took $42\,s$ with a radius of $0.06\,m$. This shows that while both models produce similar turning radii, the FED-LSTM model outperform the EF model in time efficiency by 47.6\%.

\begin{figure}[ht]
    \centering
    \includegraphics[width=1\linewidth]{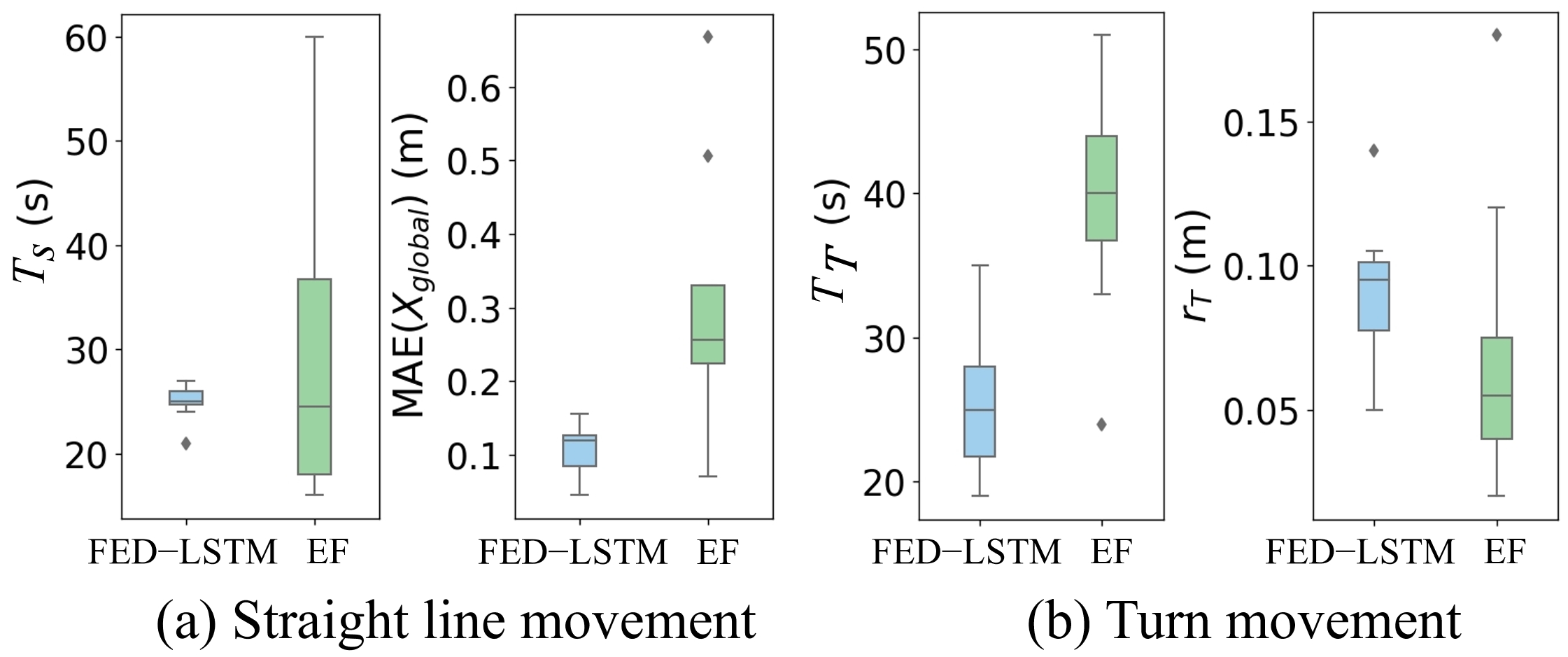} 
    \caption{Performance Comparison of Optimal Solutions: FED-LSTM vs. EF Models. The figure highlights key performance metrics (MAE ($X_{\text{global}}$), turning radius, and swimming time) for both straight-line and turning motions. Each data point represents the experimental result of one optimal solution, directly comparing the performance of these models.}
    \label{hardwaresolutions}
\end{figure}

We provide a systematic analysis of the performance of the FED-LSTM and EF models in both straight-line and turning motions, focusing on MAE($X_{\text{global}}$), swimming time, and turning radius. The following subsections present detailed comparisons.

\subsubsection{Swimming Time (Straight-Line Motion)}
As shown in Fig.~\ref{hardwaresolutions}(a) left, the swimming time of the FED-LSTM model is consistently between 21 and $27\,s$ with minimal fluctuations, demonstrating efficiency in time-sensitive tasks. In contrast, the swimming time of the EF model varies significantly, with some groups (e.g. $EF_4$ and $EF_6$) taking up to $60\,s$, while others (e.g. $EF_1$ and $EF_3$) perform similarly to the FED-LSTM model, indicating inconsistent efficiency under different conditions.

\subsubsection{Mean Absolute Error}
 MAE($X_{\text{global}}$) compares the mean absolute error between the FED-LSTM and EF models during the straight-line motion, shown in Fig.~\ref{hardwaresolutions}(a) right. The FED-LSTM model consistently shows lower and more stable errors, around $0.1\,m$, with a narrow interquartile range (IQR), indicating reliable performance across conditions. In contrast, the EF model exhibits greater variability, with some groups (e.g. $EF_6$) showing errors over $0.5\,m$. This suggests that the EF model is less stable and less precise, particularly in tasks requiring high accuracy.

\subsubsection{Swimming Time (Turning Motion)}
During turning motions, the FED-LSTM model maintains swimming time between 19 and $35\,s$ (Fig.~\ref{hardwaresolutions}(b) left), reflecting effective path planning. This consistency across experimental groups demonstrates the model's stability and efficiency, even in complex maneuvers. In contrast, the EF model shows significant variability in turning times. Some groups (e.g., $EF_3$ and $EF_8$) perform faster turns, while others (e.g., $EF_2$ and $EF_4$) take up to $50\,s$, indicating the inconsistency of the EF model in handling complex motion patterns.

\subsubsection{Turning Radius}
$r_T$ compares the turning radii of the FED-LSTM and EF models, shown in Fig.~\ref{hardwaresolutions}(b) right. The FED-LSTM model consistently maintains a turning radius between 0.05 and $0.14\,m$ with less variance, indicating smooth and stable turning performance, critical for precise control. Conversely, the EF model shows a wide range of turning radii, with some groups (e.g., $EF_8$) reaching $0.18\,m$, while others (e.g., $EF_2$) show much smaller radii near $0.02\,m$. This large variance highlights the unreliability of the EF model in executing precise turning maneuvers, making it less suitable for tasks that require consistent movement control.

\begin{table}[htbp]
\caption{Comparison of FED-LSTM and EF Models}
\begin{center}
\begin{tabular}{|c|c|c|c|c|}
\hline
\textbf{Model} & \textbf{$T_S$\textit{(s)}} & \textbf{\textit{MAE (\(X_{\text{global}}\)) (m)}} & \textbf{$T_T$\textit{(s)}} & \textbf{\textit{\(r_T\) (m)}} \\
\hline
FED-LSTM$_1$ & 21 & 0.09 & 22 & 0.08  \\
FED-LSTM$_2$ & 26 & 0.06 & 28 & 0.05  \\
FED-LSTM$_3$ & 27 & 0.16 & 35 & 0.07  \\
FED-LSTM$_4$ & 26 & 0.05 & 27 & 0.09  \\
FED-LSTM$_5$ & 25 & 0.12 & 28 & 0.14  \\
FED-LSTM$_6$ & 25 & 0.11 & 21 & 0.10  \\
FED-LSTM$_7$ & 25 & 0.13 & 19 & 0.10  \\
FED-LSTM$_8$ & 24 & 0.13 & 23 & 0.11 \\
EF$_1$   & 23 & 0.17 & 42 & 0.06  \\
EF$_2$   & 29 & 0.26 & 51 & 0.02  \\
EF$_3$   & 16 & 0.07 & 34 & 0.06  \\
EF$_4$   & 60 & 0.51 & 50 & 0.12  \\
EF$_5$   & 18 & 0.24 & 38 & 0.04  \\
EF$_6$   & 60 & 0.67 & 40 & 0.05  \\
EF$_7$   & 25 & 0.27 & 40 & 0.04  \\
EF$_8$   & 18 & 0.26 & 33 & 0.18  \\
\hline
\end{tabular}
\label{tab:comparison}
\end{center}
\end{table}

\section{Conclusion}
In this study, we develop a swimming quadrupedal robot and show that the FED-LSTM model significantly enhances real-time prediction accuracy of hydrodynamic forces on underwater quadruped robots, delivering precise, curve-based predictions even under dynamic conditions. This adaptability to changes in motion parameters, such as amplitude and frequency, sets it apart from traditional empirical formulas used for fluid flow over flat surfaces. The FED-LSTM model outperforms the EF model in accuracy and adaptability, as shown by extensive experiments, ensuring high real-time precision in managing complex underwater forces.

Beyond accurate force prediction, the FED-LSTM model excels in gait optimization. Using NSGA-II, we demonstrate reduced deflection errors in straight-line swimming and significantly decreased turning time without compromising turning radius. This adaptability underscores the robustness of FED-LSTM for controlling legged robots in environments where precision and speed are critical.

For future development, integrating real-time sensor data (e.g., IMU, depth sensor) with hydrodynamic models driven by fluid experimental data will enhance the accuracy of force predictions in complex underwater environments. By embedding this fused model into reinforcement learning frameworks, more sophisticated underwater tasks, such as inversion and path-following, can be effectively optimized and planned.

\bibliographystyle{IEEEtran}

\bibliography{main}  

\end{document}